\documentclass[runningheads]{llncs}
\usepackage{graphicx}
\usepackage{amsmath,amssymb} % define this before the line numbering.
\usepackage{color}
\usepackage[dvipsnames]{xcolor}
\usepackage{ulem}
\usepackage{lmodern}
\usepackage{hyperref}

\usepackage{color}
\usepackage{amsmath}
\usepackage{amssymb}
\usepackage{multirow, varwidth}
\usepackage{verbatim}
\makeatletter
\newif\if@restonecol
\makeatother

\usepackage[ruled,vlined]{algorithm2e}
\usepackage{xr}
\usepackage[amsmath,thmmarks]{ntheorem}

\theorembodyfont{\normalfont}
\theoremseparator{.}
\newcommand{\bm}[1]{\boldsymbol{#1}}

\DeclareRobustCommand\onedot{\futurelet\@let@token\@onedot}
\def\onedot{.\@\xspace}

\def\eg{\textit{e.g}\onedot} 
\def\ie{\textit{i.e}\onedot}

\def\etal{\textit{et al}\onedot}

\usepackage{color}

\usepackage[width=122mm,left=12mm,paperwidth=146mm,height=193mm,top=12mm,paperheight=217mm]{geometry}
\begin{document}
% \linenumbers
\pagestyle{headings}
\mainmatter

\title{A Joint Sequence Fusion Model for Video Question Answering and Retrieval} 
% Replace with your title

\titlerunning{A Joint Sequence Fusion Model for Video VQA and Retrieval}
% Replace with a meaningful short version of your title

\authorrunning{Y. Yu , J. Kim and G. Kim}
% Replace with shorter version of the author list. If there are more authors than fits a line, please use A. Author et al.

\author{Youngjae Yu \hspace{9pt} Jongseok Kim \hspace{9pt} Gunhee Kim}
 
%Please write out author names in full in the paper, i.e. full given and family names. 
%If any authors have names that can be parsed into FirstName LastName in multiple ways, please include the correct parsing, in a comment to the volume editors:
%\index{Lastnames, Firstnames}
%(Do not uncomment it, because you may introduce extra index items if you do that, we will use scripts for introducing index entries...)

\institute{Department of Computer Science and Engineering,\\
	Seoul National University, Seoul, Korea\\
	\email{ \{yj.yu,js.kim\}@vision.snu.ac.kr, gunhee@snu.ac.kr} \\
\url{http://vision.snu.ac.kr/projects/jsfusion/}
}

\maketitle
\begin{abstract}
We present an approach named JSFusion (Joint Sequence Fusion) that can measure semantic similarity between any pairs of multimodal sequence data (\eg a video clip and a language sentence).
Our multimodal matching network consists of two key components. % for a pair of sequence data, 
First, the \textit{Joint Semantic Tensor} composes a dense pairwise representation of two sequence data into a 3D tensor.
Then, the \textit{Convolutional Hierarchical Decoder} computes their similarity score by discovering hidden hierarchical matches between the two sequence modalities.  % or predicts a word as an answer to a question
Both modules leverage hierarchical attention mechanisms that learn to promote well-matched representation patterns while prune out misaligned ones in a bottom-up manner.
% Our method successfully finds hierarchical matching patterns between complex natural language query and video representation (\ie sequence of frames or sounds). 
Although the JSFusion is a universal model to be applicable to any multimodal sequence data, this work focuses on video-language tasks including multimodal retrieval and video QA.  
 % In order to show that our approach of fusion method effectively improves the performance of multiple video-language tasks,
We evaluate the JSFusion model in three retrieval and VQA tasks in LSMDC, for which our model achieves the best performance reported so far. 
We also perform multiple-choice and movie retrieval tasks for the MSR-VTT dataset, on which our approach outperforms many state-of-the-art methods. 
% Moreover, our method successfully retrieve one sequence of modality from other sequence of modality as query. (\ie Retrieve video frames with sound or vice versa). 
\end{abstract}

\section{Introduction}
\label{sec:introduction}

%------------------------------------------------------------------------------
% Figure 1: Key idea
\begin{figure*}[t]
\centering
\includegraphics[trim=0.0cm 0.2cm 0cm 0.0cm,clip,width=0.95\textwidth]{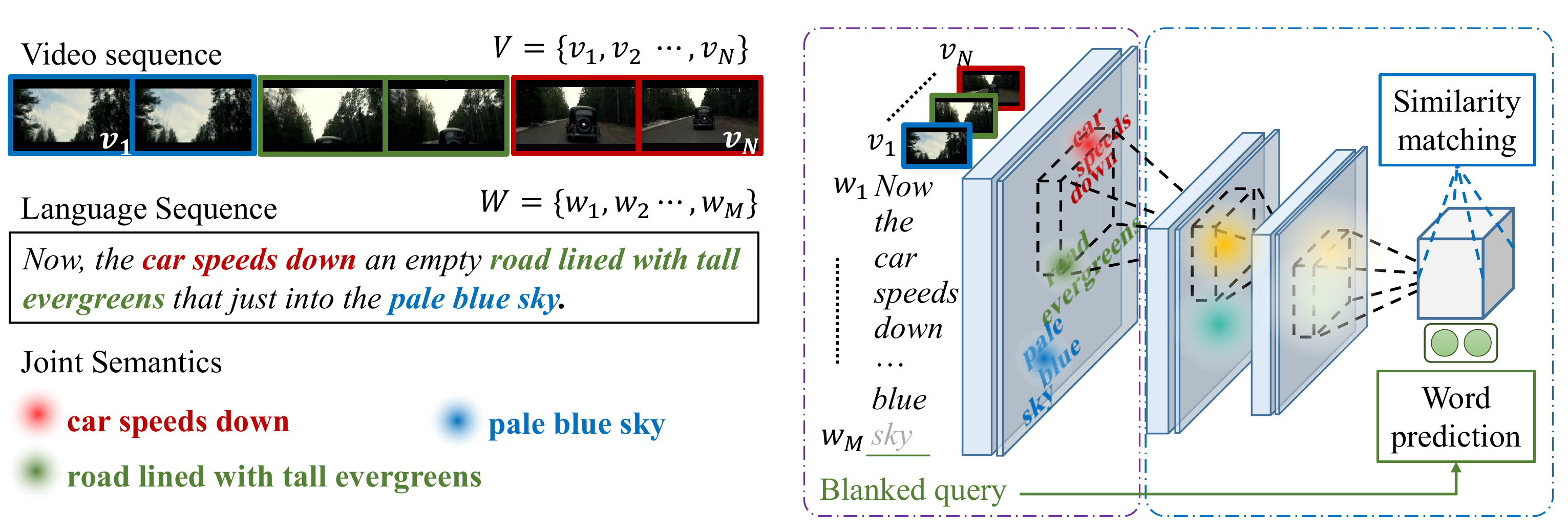}
\caption{The intuition of the Joint Sequence Fusion (JSFusion) model. 
Given a pair of a video clip and a language query, 
Joint Semantic Tensor (in purple) encodes a pairwise joint embedding between the two sequence data, 
and Convolutional Hierarchical Decoder (in blue) discovers hierarchical matching relations from JST. 
Our model is easily adaptable to many video QA and retrieval tasks.
}

% \vspace{-7pt}
\label{fig:keyidea}
\end{figure*}
%------------------------------------------------------------------------------

Recently, various video-language tasks have drawn a lot of interests in computer vision research \cite{rohrbach-arxiv-2016,xu-CVPR-2016,chen-acl-2011}, including video captioning~\cite{jeff-cvpr-2015,guadarrama-iccv-2013,rohrbach-gcpr-2015,venugopalan-iccv-2015,xu-aaai-2015,yu-cvpr-2017}, video question answering (QA)~\cite{tapaswi-cvpr-2016,jang-CVPR-2017}, and video retrieval for a natural language query~\cite{xu-aaai-2015,torabi-arxiv-2016,mayu-arxiv-2016}. 
% As demands for managing huge video content soar, discovering the relations between video data and free-form natural language queries becomes important for joint understanding for both computer vision and natural language processing. 
To solve such challenging tasks, it is important to learn a hidden join representation between word and frame sequences, for correctly measuring their semantic similarity. 
Video classification \cite{laptev-iccv-2003,laptev-cvpr-2008,soomro-crcv-2012,karpathy-CVPR-2014,caba-cvpr-2015} can be  a candidate solution, 
but tagging only a few labels to a video may be insufficient to fully relate multiple latent events in the video to a language description. % visual and sound modality.  %multimodal long sequence data such as
% There are lots of unlabelled events and relations behind video content like movie, for which inevitably, complicated annotation or queries should be provided.
Thanks to recent advance of deep representation learning, many methods for multimodal semantic embedding (\eg \cite{kiros-tacl-2014,frome-nips-2013,socher-tacl-2014}) have been proposed.
However, most of existing methods embed each of visual and language information into a single vector,
which is often insufficient especially for a video and a natural sentence. 
With single vectors for the two sequence modalities, it is hard to directly compare multiple relations between subsets of sequence data (\ie matchings between subevents in a video and short phrases in a sentence),
for which hierarchical matching is more adequate. %required to understand relationship between two sequence data.
%  in the case of videos that is often more sophisticated than images, a single video representation may not be sufficient to match with a long natural sentence. 
There have been some attempts to learn representation of hierarchical structure of natural sentences and visual scenes (\eg \cite{socher-icml-2011,socher-acl-2013} using recursive neural networks), but they require groundtruth parse trees or segmentation labels.
%  for example, using  external structure and representation of one domain such as natural language sequence or visual scene with but

In this paper, we propose an approach that can measure semantic similarity between any pairs of multimodal sequence data, by learning bottom-up recursive matches via attention mechanisms.
We apply our method to tackle several video question answering and retrieval tasks. %. 
Our approach, named as Joint Sequence Fusion (JSFusion) model, consists of two key components. %, as shown in Figure~\ref{fig:keyidea}.
First, the Joint Semantic Tensor (JST) performs dense Hadamard products between frames and words and encodes all pairwise embeddings between the two sequence data into a 3D tensor. 
JST further takes advantage of learned attentions to  refine the 3D matching tensor. 
Second, the Convolutional Hierarchical Decoder (CHD) discovers local alignments on the tensor, by using a series of attention-based decoding modules, consisting of convolutional layers and gates. 
These two attention mechanisms promote well-matched representation patterns and prune out misaligned ones in a bottom-up manner.
Finally, CHD obtains hierarchical composible representations of the two modalities, and computes a semantic matching score of the sequence pair. %  or predicts an answer word to a question.

  %such as a video sequence and a language sentence.
% We focus on developing a universal model for matching score between multimodal sequences, including consecutive image fraames, sounds and words.

% This enables our model to find hierarchical relation pattern between the targeted pair of sequences.

We evaluate the performance of our JSFusion model on multiple video question answering and retrieval tasks on LSMDC~\cite{rohrbach-arxiv-2016} and MSR-VTT~\cite{xu-CVPR-2016} datasets.  
First, we participate in three challenges of LSMDC: multiple-choice test, movie retrieval, and fill-in-the-blank,  
which require the model to correctly measure a semantic matching score between a descriptive sentence and a video clip,  
or to predict the most suitable word for a blank in a sentence for a query video.  
Our JSFusion model achieves the best accuracies reported so far with significant margins for the lsmdc tasks.  
Second, we newly create multiple-choice and movie retrieval annotations for the MSR-VTT dataset,  
on which our approach also outperforms many state-of-the-art methods in diverse video topics (\eg \textit{TV shows}, \textit{web videos}, and \textit{cartoons}).  

\begin{comment}
The LSMDC (\textit{Large Scale Movie Description Challenge}) has been one of the most active and successful challenge series to boost up the progress of video and language research.
The challenge defines four interesting tasks on the LSMDC dataset that combines two previous datasets: MPII Movie description dataset (MPII-MD)~\cite{rohrbach-cvpr-2015} and Montreal Video Annotation Dataset (M-VAD)~\cite{torabi-mvad-2015}.
We design a different base model for each of LSMDC tasks, based on JSFusion module Figure~\ref{fig:keyidea} given video and natural language sequence pair.

(i) \textit{multiple-choice test}: given a video query and five descriptive sentences, choosing the most correct one out of them,
(ii) \textit{Movie retrieval}: ranking 1,000 movie clips for a given natural language query,
and (iii) \textit{Fill-in-the-blank}: given a video and a sentence with a single blank, filling in the blank by finding a suitable word from the whole vocabulary set.
Pair of video and natural language sequence are given to solve each task. 
\end{comment}

We summarize the contributions of this work as follows.

\begin{enumerate}
\item We propose the Joint Sequence Fusion (JSFusion) model, consisting of two key components: JST and CHD. 
% although in this paper we mainly focus on language sentences and video frames. % with visuals and sounds. 
To the best of our knowledge, it is a first attempt to leverage recursively learnable attention modules for measuring semantic matching scores between multimodal sequence data.   
Specifically, we propose two different attention models, including soft attention in JST and Conv-layers and Conv-gates in CHD. 
% The Joint Semantic Tensor (JST) composes dense pairwise embeddings between two sequence data of different domain, and
% the Convolutional Hierarchical Decoder (CHD) discovers hierarchical composible representations of two sequence modalities, and computes their semantic similarity. % \yj{and composible representations}. 
\item To validate the applicability of our JSFusion model, especially on video question answering and retrieval, 
we participate in three tasks of LSMDC~\cite{rohrbach-arxiv-2016}, and achieve the best performance reported so far. 
We newly create video retrieval and QA benchmarks based on MSR-VTT~\cite{xu-CVPR-2016} dataset, on which our JSFusion outperforms many state-of-the-art VQA models.  %\yj{covering a variety of video topics} % to test language-to-video retrieval and multiple-choice QA tasks
Our source code and benchmark annotations are publicly available in our project page. %\url{http://vision.snu.ac.kr/projects/jsfusion/}.
\end{enumerate}

%%%%%%%%%%%%%%%%%%%%%%%%%%%%%%%%%%%%%%%%%%%%%%%%%%%%%%%%%%%%%%%%%%%%%%%%%%%%%%%%%%%%%%%%%%%%%%%%%%%%%%%%%%%%%%%%%%%%%%%%%%%%%%%%%%%%%%%%%%%%%%%%%%%%%%%%

\section{Related Work}
\label{sec:related_work}

Our work can be uniquely positioned in the context of two recent research directions: video retrieval and video question answering.

\textbf{Video retrieval with natural language sentences}.
Visual information retrieval with natural language queries has long been tackled via  joint visual-language embedding models~\cite{torabi-arxiv-2016,kiros-tacl-2014,hodosh-JAIR-2013,lin-CVPR-2014,vendrov-arxiv-2015,hu-cvpr-2016,mao-cvpr-2016}.
In the video domain, it is more difficult to learn latent relations between a sequence of frames and a sequence of descriptive words, given that a video is not simply a multiple of images.
Recently, there has been much progress in this line of research. Several deep video-language embedding methods \cite{xu-aaai-2015,torabi-arxiv-2016,mayu-arxiv-2016} has been developed by extending image-language embeddings~\cite{frome-nips-2013,socher-tacl-2014}. 
Other recent successful methods benefit from incorporating concept words as semantic priors~\cite{yu-cvpr-2017,yu-arxiv-2016}, or relying on strong representation of videos like RNN-FV~\cite{kaufman-iccv-2017}. 
Another dominant approach may be leveraging RNNs or their variants like LSTM to encode the whole multimodal sequences (\eg~\cite{yu-cvpr-2017,torabi-arxiv-2016,yu-arxiv-2016,kaufman-iccv-2017}).  % for  joint embedding of multimodal sequences

Compared to these existing methods, our model first finds dense pairwise embeddings between the two sequences, and then composes higher-level similarity matches from fine-grained ones in a bottom-up manner, leveraging hierarchical attention mechanisms. % to discover between multimodal sequences. 
This idea improves our model's robustness especially for local subset matching (\eg at the activity-phrase level), which places our work in a unique position with respect to previous works. % on video retrieval with free-formed nat,ural language sentences.

\textbf{Video question answering}.
% We discuss two types of Video QA, multiple-choice QA and Video Fill-In-the-Blank (FIB) QA.
VQA is a relatively new problem at the intersection of computer vision and natural language research~\cite{malinowski-nips-2014,antol-iccv-2015,goyal-cvpr-2017}.
% A similar effort is in progress in the video understanding domain where the community has quickly progressed from recent success of Visual Question and Answering.
% However, VQA dataset is difficult to collect. Furthermore, 
Video-based VQA is often recognized as a more difficult challenge than image-based one, because video VQA models must learn spatio-temporal reasoning to answer problems, which requires large-scale annotated data.
Fortunately, large-scale video QA datasets have been recently emerged from the community using crowdsourcing on various sources of data (\eg movies for MovieQA~\cite{tapaswi-cvpr-2016} and animated GIFs for TGIF-QA~\cite{jang-CVPR-2017}). 
% As the most representative of large-scale video data sets, there have been some recent efforts to create video QA datasets based on movie \textit{Descriptive Video Service} (DVS) dataset. 
Rohrbach \etal~\cite{rohrbach-arxiv-2016} extend the LSMDC movie description dataset to the VQA domain, introducing several new tasks such as multiple-choice~\cite{torabi-arxiv-2016} and fill-in-the-blank~\cite{Tegan-arxiv-2016}. 

The multiple-choice problem is, given a video query and five descriptive sentences, to choose a single best answer in the candidates. 
To tackle this problem, ranking losses on deep representation~\cite{yu-cvpr-2017,jang-CVPR-2017,torabi-arxiv-2016} or nearest neighbor search on the joint space~\cite{kaufman-iccv-2017} are exploited. 
% Common objective is retrieve correct answer in candidates and get away from negative examples. 
Torabi \etal \cite{torabi-arxiv-2016} use the temporal attention on the joint representation between the query videos and answer choice sentences. 
Yu \etal~\cite{yu-cvpr-2017} use LSTMs to sequentially feed the query and the answer embedding conditioned on detected concept words. 
% The final states of LSTMs are utilized as deep representation of answer candidates. 
% Yu \etal~\cite{yu-cvpr-2017} multiple-choice model compared in TGIF-QA benchmark with ST-VQA model~\cite{jang-CVPR-2017} to prove effectiveness of attention mechanism in video VQA. 
The fill-in-the-blank task is, given a video and a sentence with a single blank, to select a suitable word for the blank.
To encode the sentential query sentence on the video context, MergingLSTMs~\cite{mazaheri-arxiv-2016} and LR/RL LSTMs~\cite{mazaheri-iccv-2017} are proposed. 
Yu \etal\cite{yu-cvpr-2017,yu-arxiv-2016} attempt to detect semantic concept words from videos and integrate them with Bidirectional LSTM that encodes the language query.
% However, they deal with text-only fill-in-the-blank, without using video context. 
However, most previous approaches tend to focus too much on the sentence information and easily ignore visual cues. 
On the other hand, our model focuses on learning multi-level semantic similarity between videos and sentences, and consequently achieves the best results reported so far in these two QA tasks, as will be presented in section \ref{sec:experiments}.

% The proposed JSFusion model surpasses both text-only bidirectional LSTM models and simple Bidirectional LSTM method. 
% \yj{Our proposed method use joint representation from }

\section{The Joint Sequence Fusion Model}
\label{sec:jsfusion}

% We below discuss base models for four LSMDC tasks.
We first explain the preprocessing steps for describing videos and sentences in section \ref{sec:preproc}, and then discuss the two key components of our JSFusion model in section \ref{subsec:jst}--\ref{subsec:illustrative_example}, respectively. % \ref{subsec:hierarchical_decoder}
% We design a different base model for each of LSMDC tasks, while they share the joint sequence fusion architecture. 
We present the training procedure of our model in section \ref{subsec:training}, and its applications to three video-language tasks in section \ref{sec:vlmodels}. 

\subsection{Preprocessing}
\label{sec:preproc}

%------------------------------------------------------------------------------
% Figure 2 : architecture pipeline
\begin{figure*}[t]
\centering
\includegraphics[width=0.95\textwidth,trim=0cm 1.1cm 0.8cm 0cm]{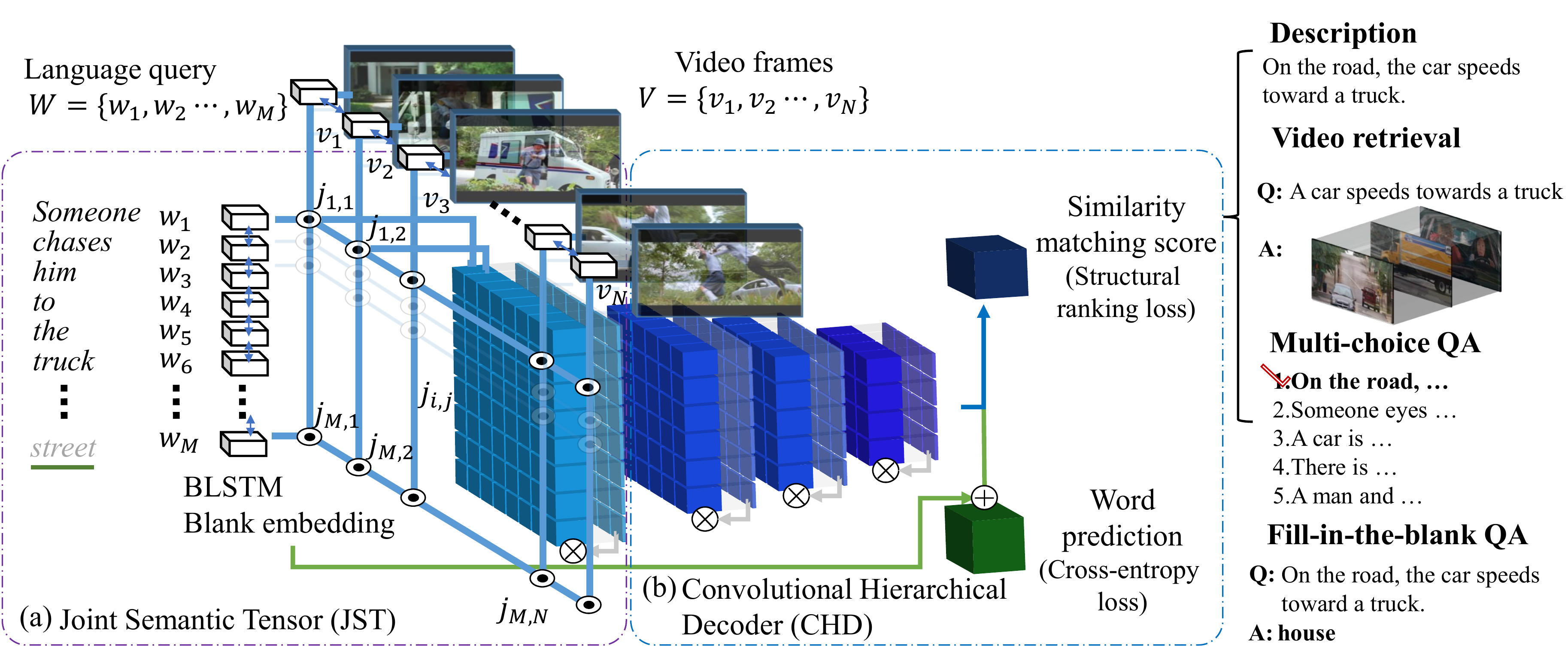}
%\vspace{-3pt}
\caption{
    The architecture of Joint Sequence Fusion (JSFusion) model. %The left part (a) of architecture is Joint Semantic Tensor (JST) and the right part (b) is Convolutional Hierarchical Decoder (CHD).
    {\bf\color{MidnightBlue}{Blue paths}} indicate the information flows for multimodal similarity matching tasks, while {\bf\color{OliveGreen}{green paths}} for the fill-in-the-blank task.
    (a) JST composes pairwise joint representation of language and video sequences into a 3D tensor, using a soft-attention mechanism. (b) CHD learns hierarchical relation patterns between the sequences, using a series of convolutional decoding module which shares parameters for each stage. $\odot$ is Hadamard product, $\oplus$ is addition, and $\otimes$ is multiplication between representation and attentions described in Eq.(\ref{eq:jst})--(\ref{eq:convatt}). We omit some fully-connected layers for visualization purpose.  
}
\label{fig:model_jsf}
%\vspace{-5pt}
 % \vspace{-7pt}

\end{figure*}
%------------------------------------------------------------------------------

\textbf{Sentence representation}.
We encode each sentence in a word level. We first define a vocabulary dictionary $\mathcal V$ by collecting the words that occur more than three times in the dataset.
(\eg the dictionary size is $|\mathcal V| = 16,824$ for LSMDC). We ignore the words that are not in the dictionary.
Next we use the pretrained glove.42B.300d \cite{Pennington-emnlp-2014} to obtain the word embedding matrix
$\mathbf E \in \mathcal R^{d \times |\mathcal V|}$ where $d=300$ is the word embedding dimension. % and $|\mathcal V|$ is the dictionary size.
% We set $d=300$ in our implementation. 
%We let the glove vector finetuned during training.
We denote the description of each sentence by $\{\mathbf w_m \}_{m=1}^{M}$ where $M$ is the number of words in the sentence. 
We limit the maximum number of words per sentence to be $M_{max}=40$. If a sentence is too long, we discard the remaining excess words, because only 0.07$\%$ of training sentences excess this limit, and no performance gain is observed for larger $M_{max}$. 
Throughout this paper, we use $m$ for denoting the word index.

\textbf{Video representation}.
We sample a video at five fps, to reduce the frame redundancy while minimizing information loss.
We employ CNNs to encode both visual and audial information in videos.
For visual description, we extract the feature map of each frame from the pool5 layer
($\mathbb R^{2,048}$)
of ResNet-152 \cite{he-arxiv-2015} pretrained on ImageNet. % ~\cite{imagenet-ijcv-2015}.
For audial information, we extract the feature map using the VGGish~\cite{Hershey-icassp-2017} followed by PCA for dimensionality reduction ($\mathbb R^{128}$).
We then concatenate both features as the video descriptor %  as video descriptors (\ie $$).
$\{\mathbf v_n \}_{n=1}^N \in \mathbb R^{2,156\times N}$ where $N$ is the number of frames in the video.
We limit the maximum number of frames to be $N_{max}=40$.
If a video is too long, we select $N_{max}$ equidistant frames. We observe no performance gain for larger $N_{max}$. % since understanding whole video is important.
We use $n$ for denoting the video frame index.

\subsection{The Joint Semantic Tensor}
\label{subsec:jst}

The Joint Semantic Tensor (JST) first composes pairwise representations between two multimodal sequences into a 3D tensor. 
Next, JST applies a self-gating mechanism to the 3D tensor to refine it as an attention map that discovers fine-grained matching between all pairwise embeddings of the two sequences while pruning out unmatched joint representations

\textbf{Sequence encoders}.
Give a pair of multimodal sequences, we first represent them using encoders.
We use \textit{bidirectional LSTM networks} (BLSTM) encoder~\cite{Schuster-ieee-1997,hochreiter-ieee-1997} for word sequence and CNN encoder for video frames.
It is often advantageous to consider both future and past contexts to represent each element in a sequence, which motivates the use of BLSTM encoders. 
$\{ \mathbf h^f_t \}_{t=1}^T$ and $\{ \mathbf h^b_t \}_{t=1}^T$ are the forward and backward hidden states of the BLSTM, respectively:
\begin{align}
\label{eq:blstm}
    \mathbf h^f_t = \mbox{LSTM} (\mathbf x_t , \mathbf h^f_{t-1}  ), \hspace{9pt}
    \mathbf h^b_t = \mbox{LSTM} (\mathbf x_t , \mathbf h^b_{t+1}  ), 
\end{align}%
where we set $\mathbf h^b_t, \mathbf h^f_t \in \mathbb R^{512}$, with initializing them as zeros: $\mathbf{h}^b_{T+1} = \mathbf{h}^f_0 = \mathbf 0$.
Finally, we obtain the representation of each modality at each step by 
concatenating the forward/backward hidden states and the input features: 
%$\mathbf x_{v,t} = [\mathbf h^f_{v,t}, \mathbf h^b_{v,t}, \mathbf v_t]$ for video frames, and 
$\mathbf x_{w,t} = [\mathbf h^f_{w,t}, \mathbf h^b_{w,t}, \mathbf w_t]$ for words.
For visual domain, we use 1-d CNN encoder representation for $v_t$, $h^{cnn} \in \mathbb R^{2,048}$ instead, $\mathbf x_{v,t} = [\mathbf h^{cnn}_{v,t}, \mathbf v_t]$.
% $\mathbf v_n, \mathbf w_m \in \mathbb R^{1,024}$. % where $n,m$ are position index of video frame or word.

%-------------------------------------------------------------------------------------------
% Table : The architecture of the composition decoder.

\begin{table}[t]
  %\footnotesize
  \setlength\tabcolsep{8.5pt} % default value: 6pt
    \centering
\begin{tabular}{|c|c|}
	\hline
	FC layers   &  size  \\ \hline    %~\cite{su:2016:ACCV}
	D$1^v$,D$1^w$   & 512    \\ 
	D2   & 512    \\ 
	D3, D4   & 512   \\ \hline
	D5   & 256   \\ 
	D6   & 256   \\ 
	D7   & 128   \\ 
	D8   & 1     \\ \hline
\end{tabular}
  %\hfill
  \begin{tabular}{|c|c|c|}
    \hline
    Conv layer       &  kernel/stride   & channel \\ \hline
    Conv1     &  3 $\times$ 3 / 1& 256    \\ 
    ConvG1     &  3 $\times$ 3 / 1& 1    \\ \hline
    Conv2     &  3 $\times$ 3 / 1& 256    \\ 
    ConvG2     &  3 $\times$ 3 / 1& 1    \\ \hline
    Conv3     &  3 $\times$ 3 / 2& 256    \\ 
    ConvG3     &  3 $\times$ 3 / 2& 1    \\ \hline
    MeanPool        &  17 $\times$ 17 / 17 & 256    \\ \hline
  \end{tabular} 
\medskip

  \caption{
    The detailed setting of layers in the JSFusion model. No padding is used for each layer. $Dk$ means a fully-connected dense layer, and $Convk$ and $ConGk$ indicate convolutional and convolutional-gating layer, respectively. %More detailed hyperparameter setting can be found in the supplementary file. % As a detailed reference, our code will be released. 
  }
% \vspace{-10pt}
\label{tbl:layer_detail}
\end{table}
%-------------------------------------------------------------------------------------------

\textbf{Attention-based joint embedding}. 
We then feed the output of the sequence encoder into fully-connected (dense) layer $[D1]$ for each modality separately, which results in $ D1^v ( \mathbf x_v ), D1^w ( \mathbf x_w ) \in \mathbb R^{d_{D1}}$, where $d_{D1}$ is a hidden dimension of $[D1]$. 
We summarize the details of all the layers in our JSFusion model in Table~\ref{tbl:layer_detail}. 
Throughout the paper, we denote fully-connected layers as $Dk$ and convolutional layers as \textit{Conv}$k$.

Next, we compute attention weights $\bm \alpha$ and representation $\bm \gamma$, from which we obtain the  JST as a joint embedding between every pair of sequential features:
\begin{align}
\label{eq:jst} 
    \mathbf j_{nm} = \alpha_{nm} \bm\gamma_{nm}, \ \mbox{where} \ \ &
    \alpha_{nm} = \sigma(\mathbf w^T  D2( \mathbf t_{nm} )) , \; \bm \gamma_{nm} = D4 ( D3 (\mathbf t_{nm}) ), \\
    &\mathbf t_{nm} = D1^v ( \mathbf x_{v,n} ) \odot {D1}^w ( \mathbf x_{w,m}).
\label{eq:att} 
\end{align}
\noindent $\odot$ is a hadamard product, $\sigma$ is a sigmoid function, and $\mathbf w \in \mathbb R^{d_{D2}}$ is a learnable parameter. 
Since the output of the sequence encoder represents each frame conditioned on the neighboring video (or each word conditioned on the whole sentence), 
the attention $\alpha$ is expected to figure out which pairs should be more weighted for the joint embedding among all possible pairs.
For example of Fig.\ref{fig:model_jsf}, expectedly, $\alpha_{3,6}(v_3, w_6) > \alpha_{8,6} (v_8, w_6)$, if $w_6$ is \textit{truck}, 
and the third video frame contains the \textit{truck} while the eighth frame does not. 

From Eq.(\ref{eq:jst})--(\ref{eq:att}), we obtain JST in a form of 3D tensor: $\mathbf J = [\mathbf j_{n,m}]^{n=1:N_{max}}_{m=1:M_{max}} $ and  $\mathbf J \in \mathbb R^{N_{max} \times M_{max} \times d_{D4}}$.

\subsection{The Convolutional Hierarchical Decoder}
\label{subsec:hierarchical_decoder}
%------------------------------------------------------------------------------
% Figure 4: Convolutional Hierarchical Decoder
\begin{figure*}[t]
\centering
\includegraphics[trim=0.0cm 0.2cm 0cm 0.0cm,clip,width=0.95\textwidth]{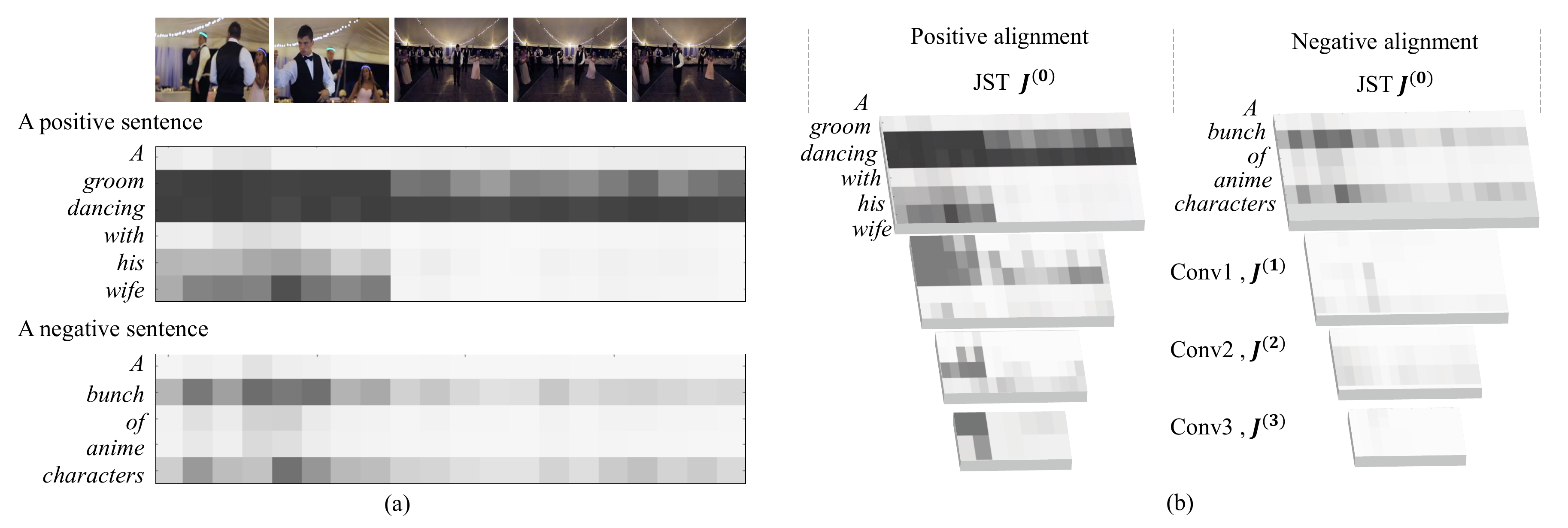}
%\vspace{-10pt}
\caption{Attention examples for (a) Joint Semantic Tensor (JST) and (b) Convolutional Hierarchical Decoder (CHD). Higher values are shown in darker.
(a) JST assigns high weights on positively aligned joint semantics in the two sequence data. Attentions are highlighted darker where words coincide well with frames. %$(n,m)$. 
(b) Each layer in CHD assigns high weights to where structure patterns are well matched between the two sequence data. 
For a wrong pair of sequences, a series of Conv-gating ($ConvG2$) prune out misaligned patterns with low weights. 
}
% \vspace{-5pt}

% \vspace{-3pt}
\label{fig:chd_att}
\end{figure*}
%------------------------------------------------------------------------------
The Convolutional Hierarchical Decoder (CHD) computes a compatibility score for a pair of multimodal sequences by exploiting the compositionality in the joint vector space of JST.
We pass the JST tensor through a series of a convolutional (Conv) layer and a Conv-gating block, whose learnable kernels progressively find matched embeddings from those of each previous layer.
That is, starting from the JST tensor, the CHD recursively activates the weights of positively aligned pairs than negatively aligned ones. 

Specifically, we apply three sets of Conv layer and Conv-gating to the JST: %, to find locally relevant meaning pairs:
\begin{align}
\label{eq:convatt}
    \mathbf J^{(k)} = Convk(\mathbf J^{(k-1)})\cdot \sigma(ConvGk(\mathbf J^{(k-1)}))
\end{align}
for $k=1,2,3$. We initialize $\mathbf J^{(0)} = \mathbf J$ from the JST, and  [$Convk$] is the $k$-th Conv layer for joint representation, [$ConvGk$] is the $k$-th Conv-gating layer for matching filters, 
whose details are summarized in Table \ref{tbl:layer_detail}.

We apply mean pooling to $\mathbf J^{(3)}$ to obtain a single  video-sentence vector representation $\mathbf J_{out}$ (\eg $\mathbb R^{17 \times 17 \times 256} \rightarrow \mathbb R^{1 \times 1 \times 256} $). 
Finally, we compute similarity matching score by feeding $\mathbf J_{out}$ into four dense layers $[D5, D6, D7, D8]$:
\begin{align}
     \mbox{score} = &\mathbf W_{D8}(D7(D6(D5(\mathbf J_{out})))) + \mathbf b_{D8} \\ \nonumber
    \mbox {where } & Dk (\mathbf x) = \mbox{tanh} (\mathbf W_{Dk} \mathbf x + \mathbf b_{Dk} ), \ \ k= 5,6,7.
\end{align}
We use the tanh activation for all dense layers except $[D8]$. % that has no activation function.

\subsection{An Illustrative Example of How the JSFusion Model Works}
\label{subsec:illustrative_example}

Fig.\ref{fig:chd_att} illustrates with actual examples how the attentions of JST and CHD work.

Fig.\ref{fig:chd_att}(a) visualizes the learned attention weights $\alpha_{nm}$ in Eq.(\ref{eq:jst}) of all pairs between frames in a video and words in a positive and a negative sentence.
The attentions are highlighted with higher values (shown in darker) when the words coincide better with the content in the frames, dominantly in a positive pair.

Fig.\ref{fig:chd_att}(b) shows the output $\mathbf J^{(k)}$ of each Conv layer and Conv-gating block in Eq.(\ref{eq:convatt}) for the same example.
During training, each Conv layer learns to compose joint embedding from the ones in the lower layer, 
while the Conv-gating layer learns frequent matching patterns in the training pairs of videos and sentences.
At test time, when it comes to compute a similarity score, the Conv-gating layers prune out misaligned patterns;
if the pair is negative where there is no common aligned structure in the two sequences, as shown in the right of Fig.\ref{fig:chd_att}(b), 
most elements of $\mathbf J^{(k)}$ have very low values. 
As a result, the CHD can selectively filter lower-layer information that needs to be propagated to the final-layer representation,
and the final layer of CHD assigns a high score only if the jointly aligned patterns are significant between the sequence data.

The motivation behind the JSFusion model is that long sequence data like videos and sentences are too complicated to compare them in a single vector space,
although most previous approaches depend on single LSTM embedding such as neural visual semantic embedding~\cite{kiros-tacl-2014} and previous LSMDC winners~\cite{kaufman-iccv-2017,yu-cvpr-2017}.
Instead, in our approach, JST first composes a dense pairwise 3D tensor representation between multimodal sequence data, from which CHD then
exploits convolutional gated layers to learn multi-stage similarity matching. %  finds multiple matches at word and frame levels and 
Therefore, our JST model can be more robust for detecting partial matching between short phrases and subhots.

% \yj{When in comes to fill-in-the-blank QA task, joint representations in lower layer are feed-forwarded through hidden structured patterns to compose higher-level joint representation to predict answer word from given query and video data pair. }

\subsection{Training}
\label{subsec:training}

We train our JSFusion model using the ranking loss.
Each training batch consists of $L$ video-sentence pairs, %in which 
including a single positive pair and $L-1$ randomly sampled negative pairs.
% for each sentence, we set one corresponding video as a positive and randomly sampled videos as negatives. 
We use batch shuffling in every training epoch.
Finally, we train the model using a max-margin structured loss objective as follows:
\begin{align}
    \mathcal{L} = & \sum_k \sum_{l=1}^{L} \max(0, S_{k,l} - S_{k,l^*} + \Delta) + \lambda ||\mathbf \theta||^2
    \label{eq:ret_maxmarginloss}
\end{align}
where $l^*$ is the answer pair among $L$ candidates, $\lambda$ is a hyperparameter and $\theta$ denotes weight parameters.
This objective encourages a positive video-sentence pair to have a higher score than a misaligned negative pair by a margin $\Delta$.
We use $\lambda = 0.0005, \Delta = 10$ in our experiments.
We train all of our models using the Adam optimizer \cite{kingma-iclr-2015}, with an initial learning rate in the range of $10^{-4}$.
For regularization, we apply batch normalization~\cite{Sergey-icml-2015} to every dense layer. %, and add dropout~\cite{srivastava-jmlr-2014} with a rate of 0.5.

\subsection{Implementation of Video-Language Models}
\label{sec:vlmodels}

We below discuss how the JSFusion model is implemented for three video-language tasks, video retrieval, multiple-choice test, and fill-in-the-blank.
We apply the same JSFusion model to both video retrieval and multiple-choice test with slightly different hyperparameter settings. 
%in Figure~\ref{fig:model_jsf}. 
For the fill-in-the-blank, we make a minor modification in our model to predict a word for a blank in the middle of the sentence. 

\begin{comment}

For multiple-choice test and movie retrieval tasks, we use similarity matching score between two multimodal sequences. 
For fill-in-the-blank task, we compute matching embedding in every possible (\ie noun, verb) blank positions and predict the correct word. 
For movie description task, we use similarity matching score from retrieval model and retrieve $k$ best plausible sentences. Among retrieved sentences, we select best one based on consensus score.  

For better understanding of our models, we outline the four LSMDC tasks as follows:
(i) \textit{Movie description}: generating a single descriptive sentence for a given movie clip,
(ii) \textit{Fill-in-the-blank}: given a video and a sentence with a single blank, finding a suitable word for the blank from the whole vocabulary set,
(iii) \textit{Multiple-choice test}: given a video query and five descriptive sentences, choosing the correct one out of them, and
(iv) \textit{Movie retrieval}: ranking 1,000 movie clips for a given natural language query.
\end{comment}

\textbf{For retrieval}.
%\label{subsec:model_retrieval}
The retrieval model takes a query sentence and ranks 1,000 test videos according to the relevance between the query and videos.
% Table~\ref{tbl:layer_archi} shows dimension we use for the Dense layers and Conv layers.
For training, we set $L=10$ as the size of each training batch. % consists of  video-sentence  pairs;
At test, for each query sentence $k$, we compute scores $\{S_{k,l}\}_{l}$ for all videos $l$ in the test set.
From the score matrix, we can rank the videos for the query.
% As will be presented in section \ref{sec:quant_results}, an ensemble of multiple score matrices
% is used for the final retrieval, which yields much better performance.
% In both single and ensembled model, 
As will be presented in section \ref{sec:quant_results} and \ref{sec:qual_results}, our method successfully finds hierarchical matching patterns between complex natural language query and video frames with sounds.

\textbf{For multiple-choice test}.
%\label{subsec:model_mc}
The multiple-choice model takes a video and five choice sentences among which only one is the correct answer.
Since our model can calculate the compatibility score between the query video and each sentence choice, we use the same model as the retrieval task.
We simply select the choice with the highest score as an answer.
% At test stage, we compute compatibility scores between given video and 5 candidate sentences, and 
For training, we set $L=10$ so that each training batch contains 10 pairs of videos and sentences, %of multiple-choice task 
which include only a single correct sentence, four wrong choices, and 5 randomly selected sentences from other training data. 
% In every epoch, we make new multiple-choice question for each videos by setting corresponding sentence as positive choice and randomly select 4 different sentences from training set as negative choices. 

\textbf{For fill-in-the-blank}.
% \label{subsec:model_blank}
The fill-in-the-blank model takes a video and a sentence with one blank, and predict a correct word for the blank. 
Since this task requires more difficult inference (\ie selecting a word out of vocabulary $\mathcal V$, instead of computing a similarity score),
we make two modifications as follows. 
First, we use deeper dimensions for layers: $d_{D1}= d_{D5} = d_{D6} = d_{D7} =1,024$, $d_{D2} = d_{D3} = d_{D4} = 2,048$,  
$\; d_{D8} = |\mathcal V|$, $\; d_{Conv1\_1} = d_{Conv2\_1} =  d_{Conv3\_1} = 1,024$, instead of the numbers in Table \ref{tbl:layer_detail}.

Second, we add a skip-connection part to our model, which is illustrated as the green paths of Figure~\ref{fig:model_jsf}. 
Letting $b$  as the blank position in the query sentence, we use the BLSTM output from the blank word token \texttt{BLANK} as a sentential context of the blank position: $\mathbf t_{b} = \mbox {D1}^w ( \mathbf w_{b})$. 
We make a summation between the output of [D7] $\in \mathbb R^{1,024}$ and the sentential context $\mathbf t_{b} \in \mathbb R^{1,024}$, 
and then feed it into [D8] to predict a word.

For training, we set the batch size as $L=32$. 
We use the different objective, the cross-entropy loss, because this task is classification rather than ranking:
\begin{align}
    \mathcal{L} = - \log p(\mathbf y) + \lambda ||\mathbf \theta||^2
    \label{eq:fib_crossentropyloss}
\end{align}%
where $\theta$ denotes weight parameters and $\lambda = 0.0005$.
We use dropout with a rate of 0.2.

%%%%%%%%%%%%%%%%%%%%%%%%%%%%%%%%%%%%%%%%%%%%%%%%%%%%%%%%%%%%%%%%%%%%%%%%%%%%%%%%%%%%%%%%%%%%%%%%%%%%%%%%%%%%%%%%%%%%%%%%%%%%%%%%%%%%%%%%%%%%%%%%%%%%%%%%
\section{Experiments}
\label{sec:experiments}

We report the experimental results of JSFusion models for the three tasks of LSMDC~\cite{rohrbach-arxiv-2016} and two tasks of MSR-VTT \cite{xu-CVPR-2016}. 
% More experimental results and implementation details can be found in the supplementary file.

%-------------------------------------------------------------------------------------------
% Table 1 : Description
\begin{table*}[tb]
\setlength\tabcolsep{6pt} % default value: 6pt
\centering
\small
\newcommand{\ranked}[1]{\xspace\scriptsize\sf{(#1)}}
\begin{tabular}{|l|cc|cc|cc|cc|}
\hline
\multicolumn{1}{|c|}{Tasks }           & \multicolumn{8}{c|}{\footnotesize Movie Retrieval} \\ \hline
\multicolumn{1}{|c|}{Metrics}           & \multicolumn{2}{c|}{\footnotesize R@1}      & \multicolumn{2}{c|}{\footnotesize R@5 } &  \multicolumn{2}{c|}{\footnotesize R@10 } & \multicolumn{2}{c|}{\footnotesize MedR } \\ \hline
\multicolumn{1}{|c|}{Dataset }           & L     & M         & L      & M        & L      & M        & L     & M         \\ \hline
% Tip: To use auto align in VIM, press "vi[ga*&" somewhere within this block [[[
LSTM-fusion                                 & 3.0   & 3.0       & 8.9    & 9.6      & 15.9   & 17.1     & 95    & 67       \\ 
SA-G+SA-FC7 \cite{torabi-arxiv-2016}        & 3.0   & 3.1       & 8.8    & 9.0      & 13.2   & 13.4     & 114   & 91       \\
LSTM+SA-FC7 \cite{torabi-arxiv-2016}    & 3.3   & 3.2       & 10.2   & 11.1     & 15.6   & 15.7     & 88    & 69        \\
C+LSTM+SA-FC7 \cite{torabi-arxiv-2016} & 4.3   & 4.2       & 12.6   & 12.9     & 18.9   & 19.9     & 98    & 55        \\
VSE-LSTM \cite{kiros-tacl-2014}        & 3.1   & 3.8        & 10.4    & 12.7    & 16.5   & 17.1      & 79    & 66         \\
EITanque \cite{kaufman-iccv-2017}      & 4.7   & 4.7       & 15.9   & 16.6     & 23.4   & 24.1     & 64    & 41        \\
SNUVL  \cite{yu-arxiv-2016}            & 3.6   & 3.5       & 14.7   & 15.9     & 23.9   & 23.8     & 50    & 44        \\ 
CT-SAN           \cite{yu-cvpr-2017}   & 4.5   & 4.4       & 14.1   & 16.6     & 20.9   & 22.3     & 67    & 35        \\ %
% CT-SAN (Ensemble) \cite{yu-cvpr-2017}  & 5.1   & 4.9       & 16.3   & 20.1     & 25.2   & 29.2     & 46    & 26        \\ 
Miech \etal \cite{miech-iccv-2017}     & 7.3   & --        & 19.2   & --       & 27.1   & --       & 52    & --        \\  \hline
JSTfc					   & 4.7   & 5.1       & 17.2   & 21.1     & 25.2   & 29.1     & 52    & 30        \\  
JSTlstm						   & 7.6   & 9.2       & 19.2    & 28.2    & 27.1   & 41.1     & 36    & 18        \\  
JSTmax                                 & 6.7   & 8.8       & 18.0     & 29.8   & 27.2   & 41.0     & 39    & 17        \\ 
JSTmean                                & 7.5   & 9.0       & 20.9     & 27.2   & 28.2   & 40.9     & 36    & 18        \\ \hline
JSFusion-noattention                   & 6.4   & 8.7       & 18.4     & 27.4   & 28.4   & 39.5     & 41    & 19        \\ 
JSFusion-noaudio                       & 9.0   & 9.2       & 20.9     & 28.3   & 32.1   & 41.3     & 39    & 17        \\ %
JSFusion                               & \textbf{9.1} & \textbf{10.2} & \textbf{21.2} & \textbf{31.2} & \textbf{34.1} & \textbf{43.2} & \textbf{36} & \textbf{13}\\ %
% JSFusion (Ensemble)                    & \textbf{10.8} & \textbf{16.7} & \textbf{14.1} & \textbf{34.5} & \textbf{36.7} & \textbf{47.8} & \textbf{29} & \textbf{10}\\ %
\hline
\end{tabular}
\medskip
\caption{
Performance comparison for the movie retrieval task using Recall@k (R@k, higher is better) and Median Rank (MedR, lower is better).
We report the results on the two datasets of LSMDC~\cite{rohrbach-arxiv-2016} (L) and MSR-VTT~\cite{xu-CVPR-2016} (M).
}
\vspace{-5pt}

\label{tbl:results_ret}
\end{table*}

%-------------------------------------------------------------------------------------------
% Table 1 : Description
\begin{table*}[tb]
\setlength\tabcolsep{4pt} % default value: 6pt
\centering
\small
\newcommand{\ranked}[1]{\xspace\scriptsize\sf{(#1)}}
\begin{tabular}{|l|cc|}
\hline
\multicolumn{1}{|c|}{Multiple-Choice }           & \multicolumn{2}{c|}{\footnotesize Accuracy} \\ \hline
\multicolumn{1}{|c|}{Dataset }         & L       & M           \\ \hline
% Tip: To use auto align in VIM, press "vi[ga*&" somewhere within this block [[[
LSTM-fusion                            & 52.8    & 38.3        \\ 
SA-G+SA-FC7 \cite{torabi-arxiv-2016}   & 55.1    & 55.8         \\
LSTM+SA-FC7 \cite{torabi-arxiv-2016}   & 56.3    & 59.1         \\
C+LSTM+SA-FC7 \cite{torabi-arxiv-2016} & 58.1    & 60.2      \\
VSE-LSTM \cite{kiros-tacl-2014}        & 63.0    & 67.3       \\ 
SNUVL  \cite{yu-arxiv-2016}            & 63.1    & 65.4      \\ 
ST-VQA-Sp.Tp \cite{jang-CVPR-2017}     & 63.5    & 66.1     \\ %
EITanque \cite{kaufman-iccv-2017}      & 63.7    & 65.5       \\
CT-SAN           \cite{yu-cvpr-2017}   & 63.8    & 66.4     \\ %
MLB \cite{Kim-iclr-2017}          & 69.0    & 76.1       \\

% CT-SAN (Ensemble) \cite{yu-cvpr-2017}  & 67.0    & 71.3       \\ 
\hline
JSTfc					   & 64.7    & 68.7       \\  
JSTlstm						   & 72.1    & 79.7       \\  
JSTmax                                 & 68.3    & 74.4       \\ 
JSTmean                                & 70.2    & 80.0        \\ \hline
JSFusion-noattention                   & 69.4    & 79.2        \\ 
JSFusion-VGG-noaudio                   & 68.7    & 75.6     \\
JSFusion-noaudio                       & 72.5    & 82.9     \\ %
JSFusion                               & \textbf{73.5} & \textbf{83.4}  \\ %
% JSFusion (Ensemble)                    & \textbf{78.1} & \textbf{85.9}  \\ %

\hline
\end{tabular}
\hfill
% Table 2: FIB
\begin{tabular}{|l|c|}
\hline
%\multicolumn{2}{|c|}{\footnotesize Fill-in-the-Blank}                  \\ \hline
Fill-in-the-Blank  & {\footnotesize Accuracy}   \\ \hline
Text-only BLSTM \cite{Tegan-arxiv-2016} & 32.0   \\
Text-only Human \cite{Tegan-arxiv-2016} & 30.2  \\
GoogleNet-2D + C3D \cite{Tegan-arxiv-2016}  & 35.7                     \\
Ask Your Neurons \cite{tzeng-iccv-2015}   & 33.2 \\
Merging-LSTM \cite{mazaheri-arxiv-2016}   & 34.2                     \\
SNUVL               \cite{yu-arxiv-2016}  & 38.0                     \\ 
CT-SAN            \cite{yu-cvpr-2017}     & 41.9                     \\
% CT-SAN   (Ensemble) \cite{yu-cvpr-2017}   & 42.7              \\ 
LR/RL LSTMs            \cite{mazaheri-iccv-2017} & 40.9       \\ 
LR/RL LSTMs (Ensemble) \cite{mazaheri-iccv-2017} & 43.5       \\ 
MLB \cite{Kim-iclr-2017} & 41.6                           \\   \hline
JSTfc						             & 42.9       \\  
JSTlstm						             & 43.7       \\  
JSTmax                   & 41.3                               \\ 
JSTmean                  & 44.2                                 \\ \hline
JSFusion-noattention      & 44.5                                 \\ 
JSFusion-VGG-noaudio     & 44.2                     \\
JSFusion-noaudio         & 45.26                    \\
JSFusion                 & \textbf{45.52}                    \\
% JSFusion (Ensemble)    & \textbf{48.15}                    \\ 
\hline
Human  \cite{Tegan-arxiv-2016}              & 68.7              \\
\hline
% ]]]
\end{tabular}
\medskip
\caption{
    \textbf{Left}:
Performance comparison for the multiple-choice test using the accuracy in percentage.
We report the results on the two datasets of LSMDC (L) and MSR-VTT (M). % ~\cite{rohrbach-arxiv-2016} ~\cite{xu-CVPR-2016}
    \textbf{Right}:
    Accuracy comparison (in percentage) for the movie fill-in-the-blank task.
}
\vspace{-5pt}
\label{tbl:results_mcfib}
\end{table*}
%-------------------------------------------------------------------------------------------

\subsection{LSMDC Dataset and Tasks}
\label{sec:lsmdc_intro}

% The LSMDC dataset is based on the two previous datasets: MPII Movie description dataset (MPII-MD)~\cite{rohrbach-cvpr-2015} and Montreal Video Annotation Dataset (M-VAD)~\cite{torabi-mvad-2015}.
%The key characteristic of the LSMDC dataset is that it takes advantage of \textit{Audio Description} (AD) and \textit{Descriptive Video Service} (DVS) resources of movies to obtain linguistic descriptions of the scenes, which are originally provided for blind or visually impaired people.
% Since they are aligned well to the videos, and transcribe key visual content of the clips by professionals, they can be valuable sources for vision and language research.
% The dataset also exploits the scripts obtained from the web for some movies.

The LSMDC 2017 consists of four video-language tasks for movie understanding and captioning, 
among which we focus on the three tasks in our experiments: movie retrieval, multiple-choice test, and fill-in-the-blank.
% because they are related to video retrieval and question and answering related benchmarks.
% All task is related to movie descriptive video service (DVS) annotation ;
The challenge provides a subset of the LSMDC dataset, % consisting of short movie clips and associated sentence descriptions.
which contains a parallel corpus of 118,114 sentences and 118,081 video clips of about 4--5 seconds long sampled from 202 movies.
We strictly follow the evaluation protocols of the challenge.
We defer more details of the dataset and challenge rules to \cite{rohrbach-arxiv-2016} and the homepage\footnote{\url{https://sites.google.com/site/describingmovies/lsmdc-2017}.}.
%\subsubsection{Movie Annotation and Retrieval}
%\label{sec:lsmdc_retrieval}
% This track measures the video retrieval performance of algorithms. The track is further divided into two tasks below.

\textbf{Multiple-choice test}. Given a video query and five candidate captions, the goal is to find the correct one for the query out of five possible choices.
% We have labeled each caption in the dataset with one or multiple activity-phrase labels.
The correct answer is the groundtruth (GT) caption and four other distractors are randomly chosen from other captions that have different activity-phrase labels from the correct answer.
The evaluation metric is the percentage of correctly answered test questions out of 10,053 public-test data.

\textbf{Movie retrieval}. 
The test set consists of 1,000 video/activity phrase pairs sampled from the LSMDC17 public-test data.
Then, the objective is, given a short query activity-phrase (\eg \textit{answering phone}), to find its corresponding video out of 1,000 test videos.
The evaluation metrics include Recall@1, Recall@5, Recall@10, and Median Rank (MedR).
The Recall@$k$ means the percentage of GT videos in the first $k$ retrieved videos,
and the MedR indicates the median rank of GT videos.
The challenge winner is determined by the metric of Recall@10.
% Each algorithm predicts $1,000\times 1,000$ pairwise ranking scores between phrases and videos, from which all the evaluation metrics are calculated.

%\subsubsection{Movie Fill-in-the-Blank}
%\label{sec:lsmdc_blank}

\textbf{Movie fill-in-the-blank}. 
This track is related to visual question answering.
The task is, given a video clip and a sentence with a blank in it, to predict a single correct word for the blank.
The test set includes 30,000 examples from 10,000 clips (\ie about 3 blanks per sentence).
The evaluation metric is the prediction accuracy (\ie the percentage of predicted words that match with GTs).

\subsection{MSR-VTT-(RET/MC) Dataset and Tasks}
\label{sec:msrvtt_intro}

The MSR-VTT~\cite{xu-CVPR-2016} is a large-scale video description dataset. 
It collects 118 videos per query of 257 popular queries, and filters manually to 7,180 videos. % obtained from a commercial video search engine, 
From the videos, it selects 10K video clips with 41.2 hours and 200K clip-sentence pairs.

Based on the MSR-VTT dataset, we newly create two video-text matching tasks: (i) multiple-choice test and (ii) video retrieval. 
The task objectives for these tasks are identical to those of corresponding tasks in the LSMDC benchmark.
To collect annotations for the two tasks, we exactly follow the protocols that are used in the LSMDC dataset, as described in \cite{torabi-arxiv-2016}.

\textbf{Multiple-choice test}: 
We generate 2,990 questions in total for the multiple-choice test, using all the test video clips of MSR-VTT. %~\cite{xu-CVPR-2016}.
For each test video, we use the associated GT caption for the correct answer, while randomly sampled descriptions from other test data for four negative choices.
% The evaluation metric is the percentage of correctly answered test questions from 2,990 multiple-choice public-test data.

\textbf{Video retrieval}:
For retrieval, we first sample 1,000 pairs of video clips and description queries from the test set of MSR-VTT
We use 1,000 as the size of the test set, following the LSMDC benchmark. 
As a result, the retrieval task is to find out the video that corresponds to the query caption out of 1000 candidates.

% We conduct manual filtering with 3 professional annotators to ensure that there is no activity-phrase label intersection with correct answer.    
% The details of 

\subsection{Quantitative Results}
\label{sec:quant_results}

Table \ref{tbl:results_ret}--\ref{tbl:results_mcfib} summarize the results of our experiments for the three video-language tasks. 
For LSMDC experiments, we report the results in the published papers and the official leaderboard of LSMDC 2017\footnote{FIB : \url{https://competitions.codalab.org/competitions/11691\#results}. \\Multichoice : \url{https://competitions.codalab.org/competitions/11491\#results}.}. 
% For LSMDC 2017, we compare with the results on the public dataset reported in the official evaluation server of LSMDC 2017 as of the CVPR2018 submission deadline
% (\ie Nov 15th, 2018 PST 23:59).
% Except award winners, LSMDC participants have no obligation to disclose their identities or used technique.
% We use the corresponding IDs in the leaderboard to denote participants. 
For MSR-VTT experiments, we run some participants of LSMDC, including SNUVL, EITanque, VSE-LSTM, ST-VQA-Sp.Tp and CT-SAN, using the source codes provided by the original authors. 
We implement the other baselines by ourselves, only except Miech \etal that require an additional person tracker, which is unavailable to use. %for MSR-VTT dataset
% Note that our method denoted by (JSFusion) is not an ensemble model but learned once. 
Other variants of our method will be discussed in details below in the ablation study. 

Table \ref{tbl:results_ret}--\ref{tbl:results_mcfib} clearly show that  our JSFusion achieves the best performance with significant margins from all the baselines over the three tasks on both datasets.
That is, the two components of our approach, JST and CHD, indeed helps measure better the semantic similarity between multimodal sequences than a wide range of state-of-the-art models, such as a multimodal embedding method  (VSE-LSTM), a spatio-temporal attention-based QA model (ST-VQA-Sp.Tp), and a language model based QA inference (Text-only BLSTM).   %  and recurrent attention model on spatio-temporal regions according to given language query
Encouragingly, the JSFusion single model outperforms even the ensemble method of runner-up (LR/RL LSTMs) in the fill-in-the-blank task.

Among baselines, multimodal low-rank bilinear attention network (MLB) \cite{Kim-iclr-2017} is competitive. 
The main differences of our model from (MLB) are two-fold.
First, JSFusion embeds both a video and a sentence to feature sequences, 
whereas (MLB) represents the sentence as a single feature.
Second, JSFusion uses the self-gating to generate fine-grained matching between all pairwise embeddings of  the two sequences, 
while (MLB) uses the attention to find a position in the visual feature space that best fits for the sentence vector.  % an attention map for
Moreover, JSFusion consistently shows better performance than (MLB) in all experiments.
% Compared to other methods, our approach shows much performance gain over text-only baseline (Text-only BLSTM) which removes JST and CHD connection to inference blanked word.  
% The results of this comparative study support that our method improves multimodal understanding for fill-in-the-blank QA module.

% Table \ref{tbl:results_ret} compares between different methods for the movie retrieval task, using the metrics of Recall@$k$ (R@k) and Median Rank (MedR) on 1,000 video/ sentence test pairs.
% The JSFusion (Ensemble) obtains the video-sentence similarity matrix using an ensemble of 15 JSFusion retrieval models with different hyperparameters. 
% We defer the details of the ensemble method to the supplementary file.
% Table \ref{tbl:results_mcfib} shows that our approach also ranks the first for the multiple-choice and the fill-in-the-blank task test on both datasets.
% As in the fill-in-the-blank, the multiple-choice task also benefits from Joint Semantic Tensor and Hierarchical Decoder.

\textbf{Ablation study}.
We conduct ablation experiments on different variants of our JSFusion model and present the results in Table \ref{tbl:results_ret}--\ref{tbl:results_mcfib}.
As one naive variant of our model, we test a simple LSTM baseline (LSTM-fusion) that only carries out the Hadamard product on a pair of final states of video and language LSTM encoders. That is, (LSTM-fusion) is our JSFusion model that has neither JST nor CHD, which are the two main contributions of our model.
We train (LSTM-fusion) in the same way as done for the JSFusion model in section~\ref{subsec:training}.
As easily expected, the performance of (LSTM-fusion) is significantly worse than our JSFusion in all the tasks.

To further validate the contribution of each component, we remove or replace key components of our model with simpler ones.
To understand the effectiveness of BLSTM encoding, we test two baselines: (JSTfc) that replaces BLSTM with fully-connected layers and (JSTlstm) that replaces BLSTM with LSTM.
% More ablation studies can be founded in the supplementary material.
(JSTmax) and (JSTmean) denote our variants that use max pooling and mean pooling, instead of the $Convk$ convolutional layers in CHD.
That is, they use fixed max/mean pooling operations instead of convolutions with learnable kernels.
These comparisons reveal that the proposed CHD is critical to improve the performance of JSFusion nontrivially on all the tasks on both datasets. 
% Interestingly, these two variants show better performance than all the baselines, 
% meaning that the weighted sum of joint embedding in JST itself is highly powerful to correctly measure the similarity between multi-modal sequences. 
% Nonetheless, 
We also compare our model with (JSFusion-noattention) that discards Conv-gating operations of CHD. 
(JSFusion-noattention) shows nontrivial performance drops as MC (acc): $4.1\%p, 4.2\%p$, RET (R@10): $5.7\%p, 3.7 \%p$ for LSMDC and MSR-VTT, respectively.
Finally, we test our model with using no audio information denoted by (JSFusion-noaudio), 
which is also much better than other baselines but only slightly worse than our original model. 

%(JSFusion) outperforms the simple single-layer LSTM/BLSTM variants with the scoring layer on top of the blank location,

%-------------------------------------------------------------------------------------------
% Figure 4 : Examples of description
\begin{figure*}[!t]
\centering
\includegraphics[trim=0.0cm 0.2cm 0cm 0.2cm,clip,width=0.9\textwidth]{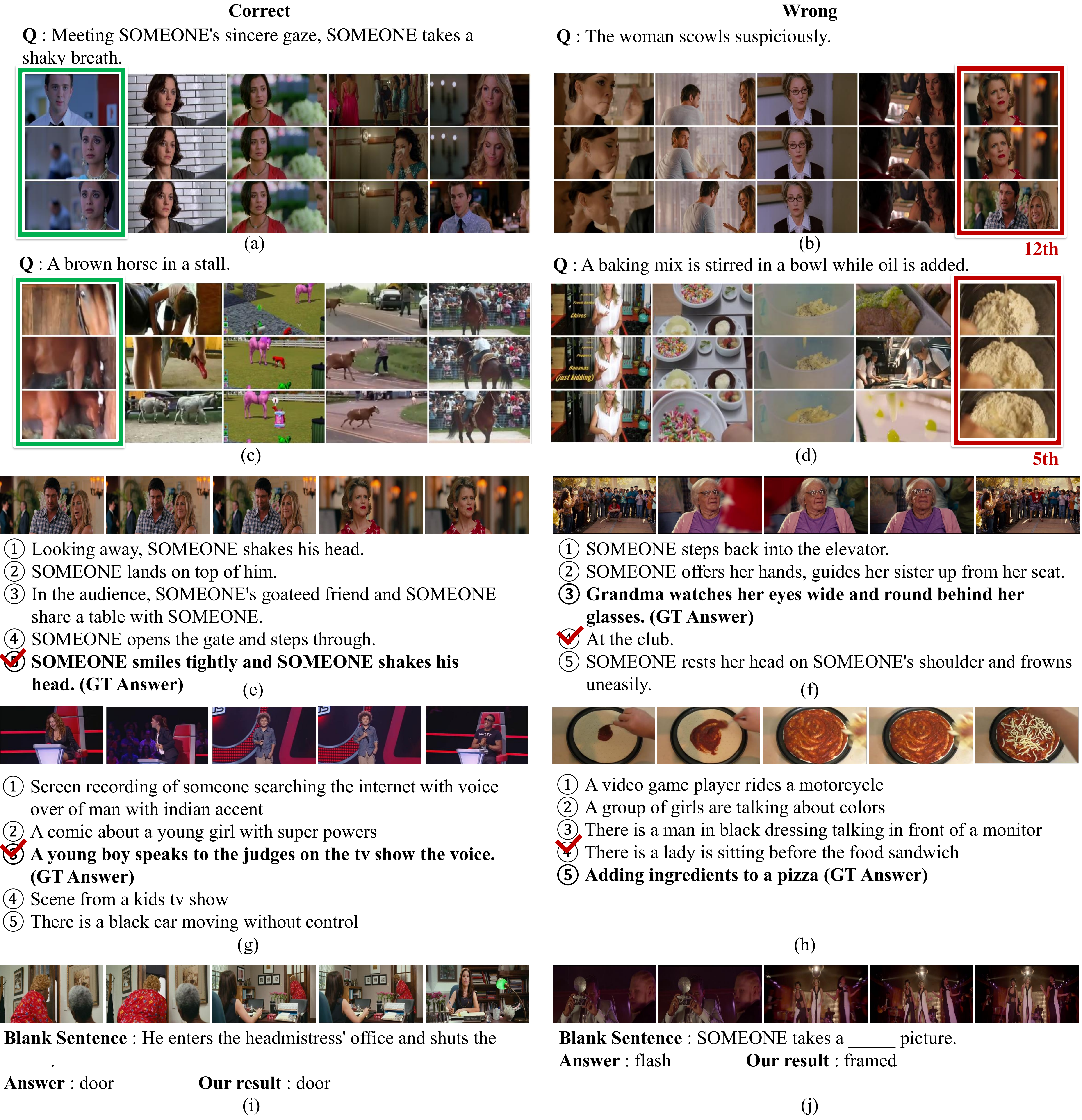}
% \vspace{3pt}
\caption{Qualitative examples of the three video-language tasks:
    movie retrieval on LSMDC (a)-(b) and MSR-VTT-RET (c)-(d), multiple-choice on LSMDC (e)-(f) and MSR-VTT-MC (g)-(h), and (i)-(j) fill-in-the-blank on LSMDC.
    The left column shows correct examples, while the right column shows near-miss examples.
    In (b),(d), we show our retrieval ranks of the GT clips (in the red box).
}
% \vspace{-6pt}

% \vspace{-10pt}
\label{fig:examples}
\end{figure*}
%-------------------------------------------------------------------------------------------

\subsection{Qualitative Results}
\label{sec:qual_results}

Fig.\ref{fig:examples} illustrates qualitative results of our JSFusion algorithm with correct (left) and near-miss (right) examples for each task.
In each set, we show natural language query and sampled frames of a video.
We present both groundtruth (GT), our prediction (Ours).

\textbf{Movie retrieval}.
Fig.\ref{fig:examples}(a) is an example that our model can understand human behaviors like \textit{gaze}. Fig.\ref{fig:examples}(b) shows the model's failure to distinguish a small motion (\eg facial expression), and simply retrieve the videos containing the face of a \textit{woman}. 
Fig.\ref{fig:examples}(c) shows that our model successfully catches the features of \textit{horses} in both web videos and 3D animation, and correctly select the highest ranking video by focusing on the word \textit{stall}.
In Fig.\ref{fig:examples}(d), although the model can retrieve relevant videos of  \textit{cooking with bowl}, it fails to find out the answer video that contains the query description of \textit{baking mix}.

\textbf{Movie multiple-choice test}.
Fig.\ref{fig:examples}(e) delivers an evidence that our model uses the whole sentence for computing matching scores, because the model successfully chooses \textcircled{\raisebox{-0.9pt}{5}} instead of \textcircled{\raisebox{-0.9pt}{1}} that shares the same phrases (\eg \textit{shakes his head}).
Fig.\ref{fig:examples}(f) is an example of focusing on a wrong video subsequence, where our model chooses the word \textit{club} by looking at a subsequence with crowded people, 
but the answer is related to another subsequence with \textit{grandmother}. 
Fig.\ref{fig:examples}(g) is an example that the model learns words in a phrase. Choice \textcircled{\raisebox{-0.9pt}{4}} can be very tempting, since it contains the word \textit{kids}, \textit{tv} and \textit{show}. But our model successfully choose the right answer by identifying that \textit{kids tv show} and \textit{kids in tv show} mean differently.
Fig.\ref{fig:examples}(h) shows that our model fails to distinguish the details.

\textbf{Movie fill-in-the-blank}.
In Fig.\ref{fig:examples}(i), the model successfully finds the answer by using both structural information of a sentence and a video (\eg \textit{door} is a likely word after \textit{shuts the}).
Fig.\ref{fig:examples}(j) is an example that the model focuses too much on the word \textit{picture} that follows the blank, instead of visual information, and  thus choose a wrong answer \textit{framed picture} rather than \textit{flash picture}.

\begin{comment}
Figure.\ref{fig:attention} illustrates visualization of attention map on Joint Semantic Tensor and subsequent layers of Convolutional Hierarchical Decoder.
We observe our model locate valid video-language joint region in attention map. 
\end{comment}

\section{Conclusion}
\label{sec:conclusion}

We proposed the Joint Sequence Fusion (JSFusion) model for measuring hierarchical semantic similarity between two multimodal sequence data.
The two key components of the model, Joint Semantic Tensor (JST) and Convolutional Hierarchical Decoder (CHD), are easily adaptable in many video-and-language tasks, including multimodal matching or video question answering. 
% Our method successfully exploit joint hidden structure between video data and natural language annotation.
We demonstrated that our method significantly improved the performance of video understanding through natural language description.
Our method achieved the best performance in challenge tracks of LSMDC, 
and outperformed many state-of-the-art models for VQA and retrieval tasks on the MSR-VTT dataset. 

Moving forward, we plan to expand the applicability of JSFusion;
since our model is usable to any multimodal sequence data, we can explore other retrieval tasks of different modalities, such as videos-to-voices or text-to-human motions.
%%%%%%%%%%%%%%%%%%%%%%%%%%%%%%%%%%%%%%%%%%%%%%%%%%%%%%%%%%%%%%%%%%%%%%%%%%%%%%%%%%%%%%%%%
\\

\textbf{Acknowledgements.} We thank Jisung Kim and Antoine Miech for helpful comments about the model. This research was supported by Brain Research Program by National Research Foundation of Korea (NRF) (2017M3C7A1047860).  Gunhee Kim is the corresponding author. %  funded by the Ministry of Science, ICT \& Future Planning

\clearpage

\bibliographystyle{splncs}
\bibliography{egbib}
\end{document}